\documentclass{article}

     \PassOptionsToPackage{numbers, compress}{natbib}
\usepackage[preprint]{neurips_2019}

\usepackage[utf8]{inputenc} % allow utf-8 input
\usepackage[T1]{fontenc}    % use 8-bit T1 fonts
\usepackage{hyperref}       % hyperlinks
\usepackage{url}            % simple URL typesetting
\usepackage{booktabs}       % professional-quality tables
\usepackage{amsfonts}       % blackboard math symbols
\usepackage{nicefrac}       % compact symbols for 1/2, etc.
\usepackage{microtype}      % microtypography
\usepackage{graphicx}
\usepackage{algorithm,amsmath,algpseudocode}

\usepackage{tablefootnote}
\graphicspath{ {./images/} }
\title{Snowball: Iterative Model Evolution and Confident Sample Discovery for Semi-Supervised Learning on Very Small Labeled Datasets}

\author{%
  Yang Li \\
\And
  Jianhe Yuan \\
\And
  Zhiqun Zhao \\
\And
  Hao Sun \\
\And
  Zhihai He \\
\AND
  Department of EECS \\
  University of Missouri \\
  Columbia, MO 65211 \\
  \texttt{\{yltb5, zzhv7, hshq7\}@mail.missouri.edu, \{yuanjia, hezhi\}@missouri.edu} \\
}

\begin{document}

\maketitle

\begin{abstract}
 In this work, we develop a joint sample discovery and iterative model evolution method for semi-supervised learning on very small labeled training sets. We propose a master-teacher-student model framework to provide multi-layer guidance during the model evolution process with multiple iterations and generations.
  The teacher model is constructed by performing an exponential moving average  of the student  models obtained from past training steps. The master network combines the knowledge of the student  and  teacher models with additional access to newly discovered samples.
   The master and teacher models are then used to guide the training of the student network by enforcing the consistence between their predictions of unlabeled samples and evolve all models when more and more samples are discovered.
Our extensive experiments demonstrate that the discovering confident samples from the unlabeled dataset, once coupled with the above master-teacher-student network evolution, can significantly improve the overall semi-supervised learning performance. For example, on the CIFAR-10 dataset, with a very small set of 250 labeled samples, our method achieves an error rate of 11.81\%, more than 38\% lower than the state-of-the-art method Mean-Teacher (49.91\%). 
   
\end{abstract}

\section{Introduction}

Deep neural networks have achieved remarkable results in many computer vision tasks such as image classification and object detection based on a  large number of labeled samples. As creating a large labeled dataset requires extensive human effort and time, semi-supervised learning provides an efficient solution for learning from large sets of unlabeled samples with small sets of labeled samples. During the past several years, it has achieved remarkable progress. 
A number of  semi-supervised learning algorithms have been developed, including regularization methods    \cite{bishop1995training, reed1992regularization, sietsma1991creating, srivastava2014dropout}, graph-based methods   \cite{zhu2003semi, Blum:2004:SLU:1015330.1015429}, entropy minimization   \cite{grandvalet2005semi}, and pseudo labeling  \cite{lee2013pseudo}. \citet{oliver2018realistic} provides a comprehensive survey of existing  semi-supervised learning methods and classify them into two major categories: \textit{consistency regularization} and \textit{entropy minimization}. 
Most recent state-of-the-art semi-supervised learning algorithms    \cite{laine2016temporal, miyato2018virtual, tarvainen2017mean} are based on the principle of consistency regularization, which learn a smooth manifold on the labeled and unlabeled samples   \cite{belkin2006manifold}. 
For example, the Mean-Teacher    \cite{tarvainen2017mean} algorithm constructs a teacher model based on exponential moving average of the student models obtained from previous training steps to guide the training of the current student model by enforcing the prediction consistency between them on unlabeled samples. These methods have demonstrated the success in training efficient models from small sets of labeled samples, such as 1000 samples (about 2\%) on the CIFAR-10 dataset. 

In this work, we propose to push this learning limit even further by developing efficient semi-supervised learning from very small sets of labeled samples. For example, on the same CIFAR-10 dataset, we can achieve successful training with only 250 labeled samples while outperforming existing methods on learning with larger sets of labeled samples. On the Street View House Number (SVHN) dataset, we can reduce the training set size from 1000 in existing literature to 100.  

To achieve this goal, we propose to explore two major ideas: (1) we extend the existing Mean-Teacher method by introducing a master-teacher-student network to provide multi-layer guidance during the model evolution process with multiple iterations and generations.
This master network combines the knowledge of the student network and  teacher network with additional access to newly discovered samples.
 Both the master and teacher models are then used to guide the training of the student network by enforcing the prediction consistence between them on unlabeled samples.
(2) We develop a confident sample discovery method and couple it with the  master-teacher-student learning to achieve continuous model evolution with more and more samples being discovered. For this reason, we refer to our method as \textbf{Snowball}. Our extensive experimental results demonstrate that our method
significantly improves the overall semi-supervised learning performance. For example, on the CIFAR-10 dataset, our method has successfully trained a model with 250 labeled samples to achieve an error rate of 11.81\%, about 38\% lower than the  state-of-the-art method Mean-Teacher (49.91\%). 

The rest of the paper is organized as follows. Section 2 reviews related work. The proposed Snowball method is presented in Section 3. Experimental results are in Section 4. Section 5 concludes the paper.

\section{Related Work}

Training semi-supervised models with a small set of labeled samples and a large set of unlabeled samples has become an important research task with significant impact in practice. The self-training approach     \cite{yarowsky1995unsupervised, riloff2003learning, rosenberg2005semi}, also called bootstrapping or self-teaching, first trains a classifier  with labeled samples and then use the pre-trained classifier to classify unlabeled samples, select the most confident unlabeled samples,  predict their labels, and use them for the next iteration of training   \cite{zhu2005semi}. Label propagation   \cite{zhu2002learning} compares unlabeled samples to labeled samples by selecting a suitable predefined distance metric and applies the label information to unlabeled dataset. Co-training   \cite{blum1998combining, mitchell1999consumer} trains two separate classifiers to learn features on two splitting datasets. Each classifier then classifies unlabeled samples and is retrained on high confident unlabeled samples with noisy labels given by the other classifier. 

Graph-based semi-supervised algorithms assume that neighbor nodes have similar labels and define a graph to represent the similarity of the samples    \cite{zhu2005semi}. \citet{blum2001learning} proposed a graph min-cut algorithm which computes the mode instead of the marginal probabilities and \citet{Blum:2004:SLU:1015330.1015429} implemented the min-cut for different graphs with perturbations. The manifold regularization method in   \cite{belkin2006manifold} applies the Reproducing Kernel Hilbert Space (RKHS) to a parameterized classifier with squared loss or hinge loss. The semi-supervised embedding algorithm   \cite{weston2012deep} modifies the regularizer by extending the classifier to embedding of data such as hidden layers, auxiliary embedding or output labels   \cite{yang2016revisiting}.

Most recent semi-supervised methods with remarkable results aim to make the model prediction to be consistent with perturbations. Transform loss  \citet{sajjadi2016regularization} and temporal ensembling   \cite{laine2016temporal} use  a similar principle when designing the  loss function. $\Pi$-model   \cite{laine2016temporal} can  be also considered as a special case of temporal ensembling by training the network and predicting likely labels for unlabeled samples over multiple epochs instead of labeling them with an outside model. The temporal ensembling model   \cite{laine2016temporal} simplifies $\Pi$-model by considering the prediction of previous training steps. The Mean-Teacher algorithm   \cite{tarvainen2017mean} uses an exponential moving average of the student models obtained from previous training steps. The student and teacher models improve each other in an iterative manner. The Smooth neighbors on teacher graph (SNTG) method    \cite{luo2018smooth} constructs a graph based on the prediction of the teacher model and forms a better teacher. Generative adversarial training  changes the training samples sightly to obtain different prediction    \cite{goodfellow2014explaining}. Virtual adversarial training (VAT)    \cite{miyato2018virtual} introduces  adversarial training for semi-supervised learning and proposes a regularization method by adding perturbation which affects the prediction of the training samples.

\section{The Proposed Snowball Method}

In the section, we present our iterative model evolution and confident sample discovery for semi-supervised learning.  

\begin{figure}
  \centering
  \includegraphics[width=0.95\linewidth]{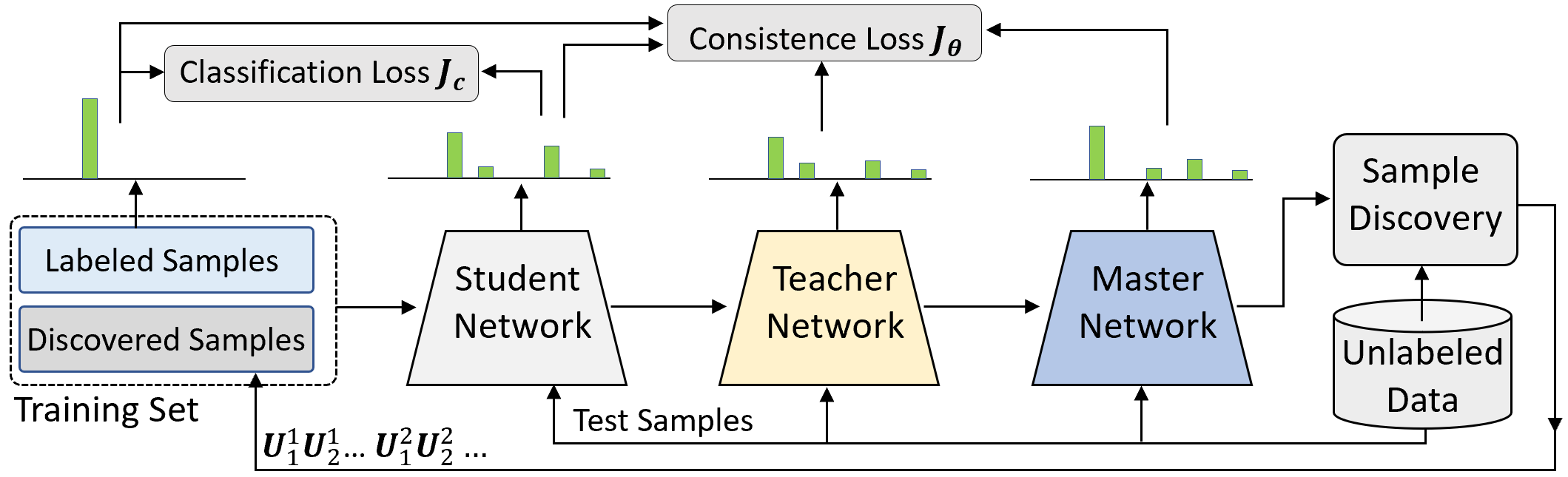}
  \caption{Overview of the proposed Snowball method.}
  \label{fig:overview}
\end{figure}

\subsection{Method Overview}

Figure \ref{fig:overview} provides an overview of the proposed Snowball method. 
In semi-supervised learning, we have access to a small set of labeled training samples, denoted by $\Omega_L$ and a large set of unlabeled samples, denoted by $\Omega_U$.
Based on $\Omega_L$, we follow the Mean-Teacher method in    \cite{tarvainen2017mean} to train the student network $\mathbf{G}_S$ guided by the teacher network $\mathbf{G}_T$. We use this successfully trained student model $\mathbf{G}_S$ to analyze unlabeled images, extract their features, compare them against the labeled images, and discover a subset of confident samples with assigned  labels. We denote this subset by $\mathbf{U}$. These discovered samples are combined with the original labeled samples to form an augmented  sample set to train a new network and form the master network model $\mathbf{G}_M$. This master network combines the knowledge of the student network $\mathbf{G}_S$ and  teacher network $\mathbf{G}_T$, as well as the knowledge of newly discovered samples $\mathbf{U}$. In Snowball, we use the master network to guide the learning of the teacher and student networks and evolve their models over multiple generations with more and more samples being discovered. Our experimental results  demonstrate that this tightly coupled model evolution and sample discovery are able to significantly improve the performance of semi-supervised learning, especially on very small sets of labeled training samples.  

\subsection{Master-Teacher-Student Network Model Evolution}

In the original Mean-Teacher method    \cite{tarvainen2017mean}, a teacher model is constructed by performing an exponential moving average (EMA) of the student network models obtained from past training steps. Each step could be multiple training epochs. This teacher model is then used to guide the training of the student network by enforcing the consistence between the predictions of unlabeled samples by the student and teacher models.

We define two terms: iteration and generation. 
In each iteration indexed by $k$, our method uses the current network model to find the additional subset of confident samples from the unlabeled samples so as to refine the models. The first iteration starts with the original labeled samples $\Omega_L$. 
One generation involves a sequence of iterations.  
In each generation indexed by $m$, we use the model obtained from the previous generation and come back to re-discover confident samples and refine models over a new sequence of iterations.  
Specifically, at generation $m$ and iteration $k$, let the corresponding student, teacher, and master network models be $\mathbf{G}_S^{m, k}$, 
$\mathbf{G}_T^{m, k}$, and 
$\mathbf{G}_M^{m, k}$ respectively. Let $\mathbf{U}_k^m$ be the set of  newly discovered samples at iteration $k$ and generation $m$.
The label for each sample in 
$\mathbf{U}_k^m$  is determined by our algorithm to be discussed in the next section.
Then, the current training set is given by 
\begin{equation}
    \Omega^{m, k}=\Omega_L\cup \mathbf{U}_1^m\cup \mathbf{U}_2^m\cup \cdots \cup \mathbf{U}_k^m.
    \label{eq:training}
\end{equation}
We use $\Omega^{m, k}$ to train and update the student, teacher, and master networks. Specifically, we first train the student network $\mathbf{G}_S^{m, k}$. Let $\mathbf{G}_S^{m, k}[t]$
be the corresponding model obtained at training step $t$. Each training step can be multiple training epochs    \cite{tarvainen2017mean}. 
At each step, the teacher model $\mathbf{G}_T^{m, k}[t]$, according to Mean-Teacher method, is constructed and updated based on the following exponential moving average:
\begin{equation}
    \mathbf{G}_T^{m, k}[t] = \alpha \cdot \mathbf{G}_T^{m, k}[t-1] + (1-\alpha)\cdot \mathbf{G}_S^{m, k}[t].
\end{equation}
The loss function for training the student network is given by 
\begin{equation}
    \mathbf{J}_S^{m, k} = \lambda_1 \cdot \mathbf{J}_C^{m, k} + \lambda_2 \cdot \mathbf{J}_\theta^{m, k}.
\end{equation}
Here, $\mathbf{J}_C^{m, k}$ represents the classification loss which is the cross-entropy between the student network prediction (softmax output vector) and the associated label over the current train set $\Omega^{m, k}$
\begin{equation}
    \mathbf{J}_C^{m, k}=\mathbb{E}_{x\in \Omega^{m, k}}\Phi\{\mathbf{G}_S^{m, k}[t](x), \mathbf{L}(x)\},
\end{equation}
where $\mathbf{L}(x) $ represents the label of the input and $\Phi\{\cdot, \cdot\}$ represents the cross-entropy. 
The consistency loss $\mathbf{J}_\theta^{m, k}$ is the cross-entropy between the student and teacher predictions
\begin{equation}
    \mathbf{J}_\theta^{m, k}=\mathbb{E}_{x\in \Omega^{m, k}}\Phi\{\mathbf{G}_T^{m, k}[t](x), \mathbf{G}_S^{m, k}[t](x)\}.
    \label{eq:cl}
\end{equation}
To construct the master network, we augment the discovered samples $\mathbf{U}_k^m$ into $\bar{\mathbf{U}}_k^m$ by additional samples, for example, 50\% more images, and use the corresponding training set formed by Eq. (\ref{eq:training}) to refine the teacher network $\bar{\mathbf{G}}_T^{m, k}[t]$. Then, the master network is obtained using the exponential moving average of the refined teacher networks
\begin{equation}
    \mathbf{G}_M^{m, k}[t] = \beta \cdot \mathbf{G}_M^{m, k}[t-1] + (1-\beta)\cdot \bar{\mathbf{G}}_T^{m, k}[t].
\end{equation}
The master network is used to guide the training of the teacher and student networks to achieve better transferability on unseen samples. To  this end, 
we augment the consistency loss in Eq. (\ref{eq:cl}) by 
\begin{equation}
    \mathbf{J}_\theta^{m, k}=\mathbb{E}_{x\in \Omega^{m, k}}\Phi\{\mathbf{G}_T^{m, k}[t](x), \mathbf{G}_S^{m, k}[t](x)\}
    + \mathbb{E}_{x\in \Omega^{m, k}}\Phi\{\mathbf{G}_M^{m, k}[t](x), \mathbf{G}_S^{m, k}[t](x)\},
    \label{eq:cl2}
\end{equation}
which has  two parts: the consistency between the teacher and student networks, and the one between the master and student networks. 

\subsection{Discovering Confident Samples}
In this work, we have found out that discovering confident samples from the unlabeled dataset, once coupled with the above master-teacher-student network evolution, can significantly improve the overall semi-supervised learning performance. Our experimental results will demonstrate that these two are tightly coupled and greatly enhance the performance of each other.
To discover confident samples and assign labels for them, we use the newly trained  master network $\mathbf{G}_M^{m, k}$ to extract the feature for each sample $x$ in the unlabeled dataset, and denote it by $F(x)$. For all samples in the labeled dataset, we also compute their features. We then compute the center for each class
\begin{equation}
    \mathbf{C}_n = \frac{1}{T_n} \sum_{x\in \Omega^{m, k},\ L(x) = n} F(x),
\end{equation}
where $T_n$ is the total number samples in class $n$ and 
$L(x)$ represents the label of $x$.
For the unlabeled image $x$, we find class center $C_{n^*}$ which has the minimum distance to $F(x)$, and then assign its label as $n^*$. In our experiments, 
we find out that samples with smaller distance have higher probability to have correct labels. In  iteration (m, k), we select the top $N^{m, k}$ samples with the smallest feature distance to their centers to form the newly discovered sample set $\mathbf{U}_k^m$.

\section{Experimental Results}
\label{gen_inst}

To evaluate the performance  of our proposed method, we use two benchmark datasets: SVHN and CIFAR-10. Following the common semi-supervised classification practice   \cite{rasmus2015semi, laine2016temporal, tarvainen2017mean, luo2018smooth}, we split the training samples between a labeled and unlabeled dataset and randomly choose 250, 1000 2000 samples from for the SVHN and CIFAR-10 datasets and compare the performance with state-of-the-art methods. To test the performance of our Snowball method on very small sets of labeled samples, we reduce the training set size to 500, 250, and 100, and compare with the Mean-Teacher method. In this scenario, we choose Mean-Teacher because it has the code publicly available and we have managed to run the code to achieve the same performance as claimed in the original paper.
We use the source code published on Github by Mean-Teacher    \cite{tarvainen2017mean} and use the same augmentation, training steps, ramp-up and EMA decay rate parameters.
But, in the original paper, the authors did not present training performance on these very small training sets.
The results are averaged over multiple runs with different random seeds. We follow the random sample strategy of Mean-Teacher and ensure that each class has the same number of labeled samples. In both the Mean-Teacher and our Snowball methods, two network structures are used: a 13-layer convolutional network (ConvNet-13) and a 26-layer Residual Network   \cite{he2016deep} (Resnet-26)
with Shake-Shake regularization   \cite{sajjadi2016regularization}.
In the following experiments, we report the error rates. 

\subsection{Performance Comparison with Existing Methods on the Same Training Sets}

\textbf{Performance Comparison on the SVHN Dataset.}
The street view house numbers (SVHN) dataset consists of 32$\times$32 pixel RGB images and these images belong to 10 classes. There are 73,257 labeled samples for training and 26,032 for testing. It has been used as the benchmark dataset for testing semi-supervised learning and previous state-of-the-art methods have already achieved low error rates which are very close to supervised learning with the full training set (73257 images).  All of the labeled and unlabeled training datasets are normalized to have zero mean and unit variance. One labeled sample and 99 unlabeled samples are assigned to each mini-batch. Table \ref{sample-table-1} shows our results on the SVHN with 250 labels. We can see that our method outperforms existing state-of-the-art methods, reducing the error rate of the second best (4.29\%) further to 3.26\%. 

\begin{table}
  \caption{Error rate percentage of ConvNet and Resnet on SVHN compared to the state-of-the-art.}
  \label{sample-table-1}
  \centering
  \begin{tabular}{lllll}
    \toprule                                   
    %\multicolumn{2}{c}{Part}                   \\
    %\cmidrule(r){1-2}
                            & 250 Labels       & 73257 Labels \\
    Method                        & 73257 Images    & 73257 Images \\
    \midrule 
    Supervised-only    \cite{tarvainen2017mean}     & 27.77 $\pm$ 3.18\%   &2.75 $\pm$ 0.10\%\\
    $\Pi$ Model    \cite{laine2016temporal}         & 9.69 $\pm$ 0.92\%  &2.50 $\pm$ 0.07\%\\
    Temporal Ensembling (TempEns)    \cite{laine2016temporal} & 12.62 $\pm$ 2.91\%   & 2.74 $   \pm$ 0.06\%\\
    %$\Pi$+SNTG    \cite{luo2018smooth}              & 5.07 $\pm$ 0.25\%  &\textbf{2.42 $\pm$ 0.05\%} \\

    %TempEns+SNTG    \cite{luo2018smooth}             & 5.36 $\pm$ 0.57\%    & 2.44 $\pm$ 0.03\%\\
    Mean-Teacher + ConvNet13     \cite{tarvainen2017mean}    & 4.35 $\pm$ 0.50\%    & 2.50 $\pm$ 0.05\%\\
    % MT+SNTG    \cite{luo2018smooth}                 & 4.29 $\pm$ 0.23\%    & 2.42 $\pm$ 0.06\%\\
    SNTG    \cite{luo2018smooth}                 & 4.29 $\pm$ 0.23\%    & 2.42 $\pm$ 0.06\%\\
    \midrule
    
    Our  Method + ConvNet13         & 4.07 $\pm$ 0.17\%   & 2.50 $\pm$ 0.05\%\\
    Our  Method + Resnet                  & \textbf{3.26 $\pm$ 0.02\%} & $-$\\
    \bottomrule
  \end{tabular}
\end{table}

\textbf{Performance Comparison on the CIFAR-10 Dataset.}
CIFAR-10 is another semi-supervised learning benchmark dataset consisting of 32$\times$32 from 10 classes. There are 50,000 labeled training samples and 10,000 testing samples. Table \ref{sample-table-2} shows the model error rates for 1000, 2000, and all 50000 training samples achieved by our method and existing methods. We can see that our method with ResNet achieves the best performance, with an error rate of 7.82\% much lower than the Mean-Teacher 10.08\% which uses the same network configurations. 
We also provide results of our method with ConvNet-13 and other methods which use similar network configurations as the ConvNet-13. We include the results for the full training set to demonstrate that all methods are having a similar starting point. In the original paper, the Mean-Teacher did not provide result with Resnet. Our method will be the same as the Mean-Teacher method is the full training set is used since the master network will never be activated.

\begin{table}

  \caption{Error rate percentage of ConvNet and Resnet on CIFAR-10 compared to the state-of-the-art.}
  \label{sample-table-2}
  \centering
  \begin{tabular}{lllll}
    \toprule                                   
    %\multicolumn{2}{c}{Part}                   \\
    %\cmidrule(r){1-2}
                        & 1000 Labels       & 2000 Labels        & 50000 Labels \\
    Method              & 50000 Images      & 50000 Images       & 50000 Images \\
    \midrule 
    Supervised-only    \cite{tarvainen2017mean}     &46.43  $\pm$ 1.21\%   &33.94 $\pm$ 0.73\%    &5.82 $\pm$ 0.15\%\\
    $\Pi$ Model    \cite{laine2016temporal}         &27.36 $\pm$ 1.20\%  &18.02 $\pm$ 0.60\%   &6.06 $\pm$ 0.11\%\\
    %$\Pi$+SNTG    \cite{luo2018smooth}              &21.23 $\pm$ 1.27\%   &14.65 $\pm$ 0.31\%    &5.19 $\pm$ 0.14\%\\
    %TempEns+SNTG    \cite{luo2018smooth}            &18.41 $\pm$ 0.52\%  &13.64 $\pm$ 0.32\%   &5.20 $\pm$ 0.14\%\\
    SNTG    \cite{luo2018smooth}            &18.41 $\pm$ 0.52\%  &13.64 $\pm$ 0.32\%   &5.20 $\pm$ 0.14\%\\
    Mean-Teacher + ConvNet13    \cite{tarvainen2017mean}        &21.55 $\pm$ 1.48\%  &15.73 $\pm$ 0.31\%    &5.94 $\pm$ 0.05\%\\
    Mean-Teacher + Resnet    \cite{tarvainen2017mean} &10.08 $\pm$ 0.41\%   &8.06 $\pm$ 0.14\%    & $-$         \\
    \midrule
    %MT (ours)                         &20.27 $\pm$ 0.06   &  15.65 $\pm$ 0.32                  &\\
    Our  Method + ConvNet13  & 17.79 $\pm$ 0.11\%  & 14.56 $\pm$ 0.38\%    &5.94 $\pm$ 0.05\% \\
    %$MT Resnet (ours)                  & 9.47 $\pm$ 0.85     &  8.20                  &\\
    Our  Method + Resnet  & \textbf{7.82 $\pm$ 0.08\%}  & \textbf{7.15 $\pm$ 0.17\%} & $-$ \\
    
    \bottomrule

  \end{tabular}
\end{table}

\subsection{Performance Comparison on Very Small Training Sets}
In the following experiments, we demonstrate the performance of our method on very small training sets and provide comparison with the Mean-Teacher method which achieves the state-of-the-art performance.  Table \ref{sample-table-3} shows the results on the CIFAR-10 dataset with the size of training set reduced from 1000 to 500 and 250. We also copy over the results of 1000 and 2000 from Table \ref{sample-table-2} for the convenience of comparison. We can see that, on very small training sets, our method significantly outperforms Mean-Teacher, reducing the error rate from 49.91\% to 11.81\% with the Resnet network. For the Convnet-13 network, the error rate is reduced from 51.79\% to 37.65\%. This 38\% performance improvement is very significant.
Table \ref{sample-table-4} shows the results on the SVHN dataset. We reduce the original training set size from 250 samples to 100 samples. We can see that our method significantly outperforms the Mean-Teacher method, reducing the error rate 15.29\% to 6.04\%.

\begin{table}
  \caption{Performance comparison with Mean-Teacher on very small training sets on CIFAR-10.}
  \label{sample-table-3}
  \centering
  \begin{tabular}{@{} p{3.9cm}p{2.15cm}p{2.15cm}|p{2.15cm}p{2.15cm} @{}}
    \toprule                                   
    %\multicolumn{2}{c}{Part}                   \\
    %\cmidrule(r){1-2}
                & 250 Labels    & 500 Labels   & 1000 Labels & 2000 Labels\\
     Method           & 50000 Images  & 50000 Images  & 50000 Images & 50000 Images\\
    \midrule
    Mean-Teacher + ConvNet13       & 51.79 $\pm$ 2.13\%   & 33.02 $\pm$ 1.60\%  & 21.55 $\pm$ 1.48\%   & 15.73 $\pm$ 0.31\% \\
    Mean-Teacher + Resnet & 49.91 $\pm$ 9.38\% & 15.87 $\pm$ 0.10\% & 10.08 $\pm$ 0.41\% & 8.06 $\pm$ 0.14\%\\
    \hline
    Our  Method + ConvNet13 & 37.65 $\pm$ 2.49\% & 22.30 $\pm$ 1.48\% & 17.79 $\pm$ 0.11\%   & 14.56 $\pm$ 0.38\%\\
    Our  Method + Resnet  & \textbf{11.81 $\pm$ 0.04\%}   & \textbf{9.15 $\pm$ 0.82\%} & \textbf{7.82 $\pm$ 0.08\%} & \textbf{7.15 $\pm$ 0.17\% } \\
    \bottomrule
  \end{tabular}
\end{table}

\begin{table}
  \caption{Performance comparison with Mean-Teacher on very small training sets on the SVHN.}
  \label{sample-table-4}
  \centering
  \begin{tabular}{@{} p{4.2cm}p{2.5cm}|p{2.5cm} @{}}
    \toprule                                   
                & 100 Labels    & 250 Labels \\
     Method   & 73257 Images  & 73257 Images \\
    \midrule
    Mean-Teacher + ConvNet13       & 46.50 $\pm$ 10.12\%   & 4.35 $\pm$ 0.50\%    \\
    Mean-Teacher + Resnet & 15.29 $\pm$ 2.63\%  & 3.53 $\pm$ 0.02\%  \\
    \hline
   Our  Method + ConvNet13 & 14.20 $\pm$ 0.59\%   & 4.07 $\pm$ 0.17\%      \\    
    Our Method + Resnet  & \textbf{6.04 $\pm$ 0.43\%}   &  \textbf{3.26 $\pm$ 0.02\%} \\
    \bottomrule
  \end{tabular}
\end{table}

\subsection{Ablation Studies and Further Algorithm Analysis}

\begin{figure}
  \centering
  \includegraphics[width=\linewidth]{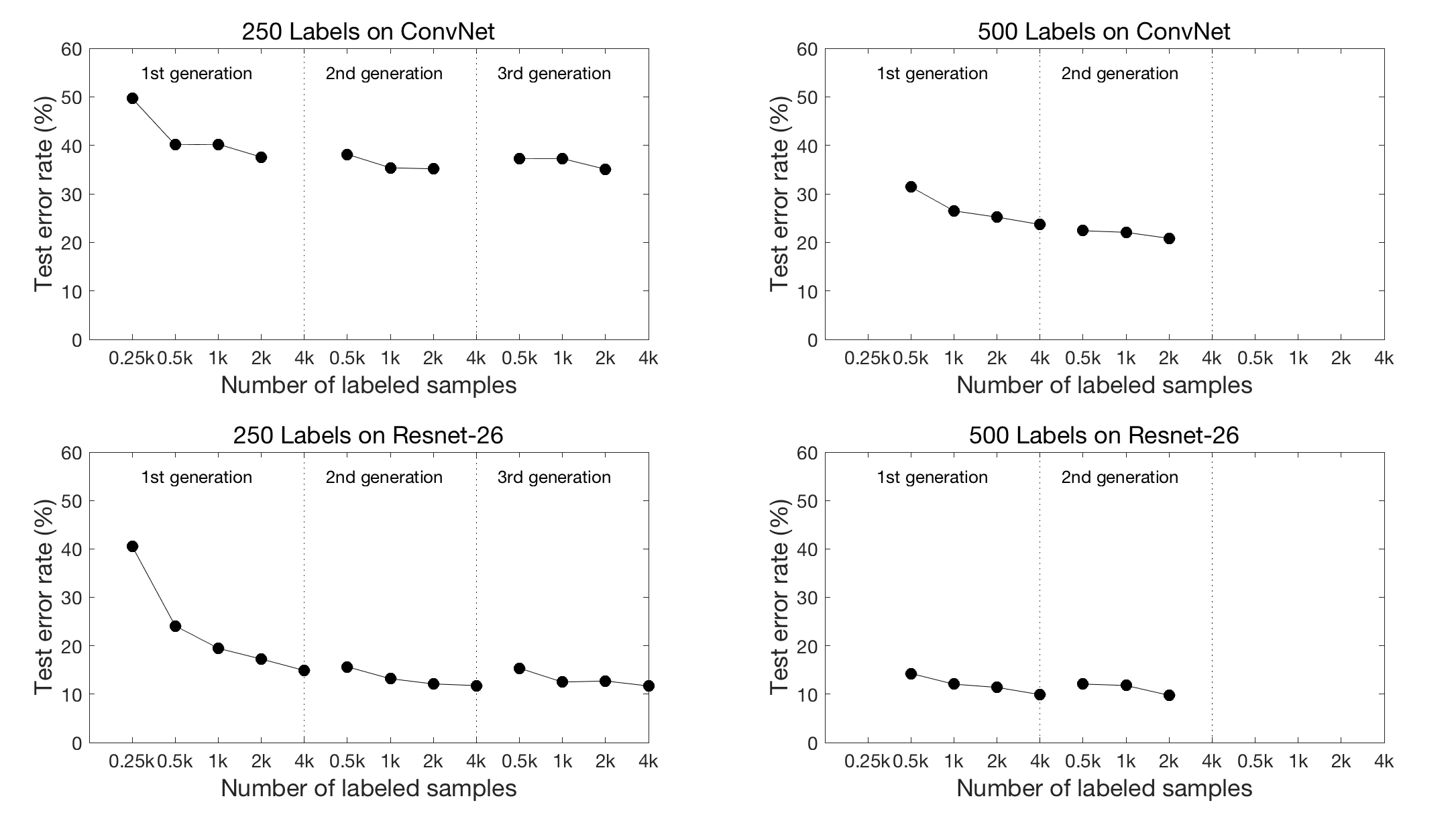}
  \caption{Error rate of iterations and generations on CIFAR-10.}
  \label{fig-converg}
\end{figure}

\textbf{A. Convergence Behaviors of Snowball}. In this experiment, we demonstrate that the proposed Snowball algorithm converges as more and more confident samples are discovered and the master-teacher-student network evolves over iterations and generations. In our experiments, we use 3 generations and each generation has 3-4 iterations. Figure \ref{fig-converg} shows the decreasing of the error rate of our method on the CIFAR-10 dataset with two network configurations, ConvNet-13 and Resnet-26, with 250 and 500 training samples. Within each generation, we use the model obtained from the previous generation to re-discover confident samples in each iteration to grow the training samples from 250, to 500, 1000, 2000, and 4000. We can see that the model is becoming more and more accurate over each iteration and generation.

\begin{figure}
  \centering
  \includegraphics[width=\linewidth]{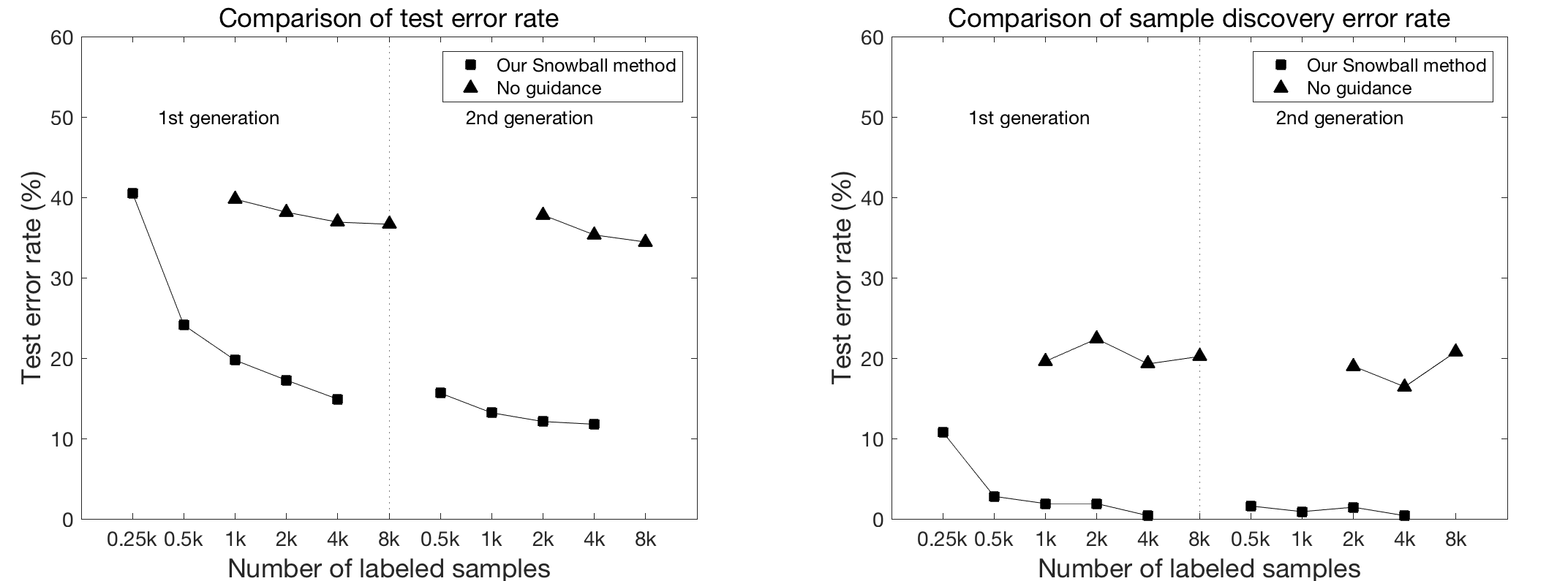}
%   \fbox{\rule[-.5cm]{0cm}{4cm} \rule[-.5cm]{4cm}{0cm}}
  \caption{Comparison of our Snowball method with self-learning without guidance by master-teacher-student networks.}
  \label{fig-guidance}
\end{figure}

% \begin{table}
%   \caption{Analysis of sample discovery}
%   \label{sample-table-distance}
%   \centering
%   \begin{tabular}{l|l|l}
%     \toprule                                   
%     %\multicolumn{2}{c}{Part}                   \\
%     %\cmidrule(r){1-2}

%     Number of Labels   & With Ground-truth Label      & Sample Error Rate \\
%     \midrule
%     500 Labels       & 14.63\%       & $-$   \\
%     500 $+$ 500 (Min)    & 12.06\%       & \textbf{1.60\%}  \\
%     500 $+$ 500 (Random) & \textbf{10.22\%}       & 16.60\% \\
%     500 $+$ 500 (Max)    & 20.73\%       & 72.60\% \\

%     \bottomrule
%   \end{tabular}
% \end{table}

\begin{table}
  \caption{Analysis of sample discovery}
  \label{sample-table-distance}
  \centering
  \begin{tabular}{l|l|l}
    \toprule                                   
    %\multicolumn{2}{c}{Part}                   \\
    %\cmidrule(r){1-2}

    Number of Labels   & With Ground-truth Label      & Sample Error Rate \\
    \midrule
    500 Labels              & 14.44\%          & $-$      \\
    500 $+$ 500 (Min)       & 10.89\%          & \textbf{0.60\%}  \\
    500 $+$ 500 (Random)    & \textbf{9.95\%}  & 14.20\% \\
    500 $+$ 500 (Max)       & 19.87\%          & 69.20\% \\

    \bottomrule
  \end{tabular}
\end{table}

\begin{table}
  \caption{Fusion methods error rate of sample discovery.}
  \label{sample-table-metric}
  \centering
  \begin{tabular}{llllll}
    \toprule                                   
    %\multicolumn{2}{c}{Part}                   \\
    %\cmidrule(r){1-2}

    Fusion methods   & Average Distance  &Feature Cascade  &Average Sorting Score  \\
    \midrule
    Noisy labels ratio   & 2.9\% & 2.7\% & 2.8\%   \\
    %                    &1.4\% &1.1\% & 1.95\% &3.95\% &7.18
    %                  &7    &11   &78 & 395 & 1436  \\
    \bottomrule
  \end{tabular}
\end{table}

% \begin{table}
%   \caption{Variation of sample discovery error rate.}
%   \label{sample-table}
%   \centering
%   \begin{tabular}{llllll}
%     \toprule                                   
%     %\multicolumn{2}{c}{Part}                   \\
%     %\cmidrule(r){1-2}

%     Number of discovered samples   &0.5k  &1k  &4k & 10k & 20k  \\
%     \midrule
%     Noisy labels ratio   & 1.4\% & 1.1\% & 1.95\% & 3.95\% & 7.18\%  \\
%     %                    &1.4\% &1.1\% & 1.95\% &3.95\% &7.18
%     %                  &7    &11   &78 & 395 & 1436  \\
%     \bottomrule
%   \end{tabular}
% \end{table}

\textbf{B. Tight Coupling between the Master-Teacher-Student Model Evolution and Confident Sample Discovery.}
In this work, we recognize that the confident sample discovering and master-teacher-student model evolution are tightly coupled. As discussed in Section 1, discovering confident samples has already been used in self-learning or bootstrap-based semi-supervised learning. Its performance is very limited. But, once combined with the master-teacher-student model evolution, we can achieve significantly improved performance. This is because the newly discovered samples have labeling errors. These errors will degrade the learning performance. But, with the master-teacher-student consistency regulation, the learning process becomes much more robust against these label errors.
Figure \ref{fig-guidance}(left) shows the comparison of our Snowball method against the self-learning methods with confident sample discovery but without guidance by the master-teacher-student network. We can see that the error rate is dramatically decreased with the master-teacher-student guidance.
In Figure \ref{fig-guidance}(right), we also show the labeling error rate in the newly discovered samples in both methods. We can see that the new samples discovered by our Snowball method is much lower, which results in significantly improved learning performance.

\textbf{C. Impact of Different Selecting Methods for Confident Sample Discovery}. 
In our Snowball new confident sample discovery method, we find the center of labeled samples which has the minimum distance to the current sample and then assign the corresponding label to this unlabeled sample. We choose the top $N$ samples with the smallest distance as the newly discovered samples for the next iteration.
On the other hand, we recognize that samples with minimum distance to existing labeled samples might be too similar to existing training samples and the contribution to the model learning and transferability will be degraded. Other possible choices include using the maximum distance criteria, or we  randomly select the top $N$ samples. 
Table \ref{sample-table-distance} shows the error rate results on CIFAR-10 dataset with 500 training samples. We need to select the next 500 new samples. The second column shows the model error rates with 500 new samples selected by three different methods, but using  the ground-truth labels. We can see that the random method achieves the best since its samples have the largest diversity. But, in practice, we do not have this ground-truth label. In this case, our minimum distance method achieves the best performance since its percentage of wrong labels in newly discovered samples is much smaller than the other two methods, as shown in the second column. This is the reason why we choose the minimum distance method in our Snowball method.

% \begin{figure}
%   \centering
%   \includegraphics[scale=0.20]{images/sampleDiscoveryErrorRate.png}
%   %\fbox{\rule[-.5cm]{0cm}{4cm} \rule[-.5cm]{4cm}{0cm}}
%   \caption{Variation of sample discovery error rate.}
% \end{figure}

\textbf{D. Different Feature Fusion Methods  for Confident Sample Discovery}

In our current method, when we identify new confident samples for automated labeling, we use the master network to extract its feature and evaluate its feature distance to existing labeled samples. 
In the following experiment, we explore additional options for the feature distance. 
For example, we can use three master network models of the past iterations to extract three separate features. 
We then fuse these features together to form a joint feature distance metric. The following three fused feature distance metrics are considered.
\textbf{(1) Average distance} - 
We use each of three features to compute the distance and then use the average of them as the distance metric for this unlabeled sample.
\textbf{(2) Feature cascade} - 
These three features are cascaded together into one combined feature vector for this unlabeled sample. We then use this cascaded feature to measure the distance to assign labels.
\textbf{(3) Average sorting score} - We use each of these three features to compute the minimum distance, then sort the samples according to their distance from the smallest to the largest. For each of these three features, we have three separate sorting scores (sorting indices), we then compute their average sorting scores and use this as the distance metric. 
Table \ref{sample-table-metric} shows the error rates for these three fusion methods on the CIFAR-10 data with 1000 training samples.  We can see that the feature cascade method has the best performance. But, the difference between these three are relatively small.

\section{Conclusion}

In this work, we have successfully developed a joint sample discovery and iterative model evolution method for semi-supervised learning from a very small labeled training set. We have established  a master-teacher-student model framework to provide multi-layer guidance during the model evolution process with multiple iterations and generations.
Both the master and teacher models are  used to guide the training of the student network by enforcing the consistence between the predictions of unlabeled samples between them and evolve all models when more and more samples are discovered.
Our extensive experiments demonstrate that the discovering confident samples from the unlabeled dataset, once coupled with the above master-teacher-student network evolution, can significantly improve the overall semi-supervised learning performance. For example, on the CIFAR-10 dataset, our method has successfully trained a model with 250 labeled samples to achieve an error rate of 11.81\%, more than 38\% lower than the state-of-the-art method Mean-Teacher (49.91\%).

\medskip

\small

\bibliography{neurips_2019}

\begin{thebibliography}{28}
\providecommand{\natexlab}[1]{#1}
\providecommand{\url}[1]{\texttt{#1}}
\expandafter\ifx\csname urlstyle\endcsname\relax
  \providecommand{\doi}[1]{doi: #1}\else
  \providecommand{\doi}{doi: \begingroup \urlstyle{rm}\Url}\fi

\bibitem[Belkin et~al.(2006)Belkin, Niyogi, and Sindhwani]{belkin2006manifold}
M.~Belkin, P.~Niyogi, and V.~Sindhwani.
\newblock Manifold regularization: A geometric framework for learning from
  labeled and unlabeled examples.
\newblock \emph{Journal of machine learning research}, 7\penalty0
  (Nov):\penalty0 2399--2434, 2006.

\bibitem[Bishop(1995)]{bishop1995training}
C.~M. Bishop.
\newblock Training with noise is equivalent to tikhonov regularization.
\newblock \emph{Neural computation}, 7\penalty0 (1):\penalty0 108--116, 1995.

\bibitem[Blum and Chawla(2001)]{blum2001learning}
A.~Blum and S.~Chawla.
\newblock Learning from labeled and unlabeled data using graph mincuts.
\newblock 2001.

\bibitem[Blum and Mitchell(1998)]{blum1998combining}
A.~Blum and T.~Mitchell.
\newblock Combining labeled and unlabeled data with co-training.
\newblock In \emph{Proceedings of the eleventh annual conference on
  Computational learning theory}, pages 92--100. ACM, 1998.

\bibitem[Blum et~al.(2004)Blum, Lafferty, Rwebangira, and
  Reddy]{Blum:2004:SLU:1015330.1015429}
A.~Blum, J.~Lafferty, M.~R. Rwebangira, and R.~Reddy.
\newblock Semi-supervised learning using randomized mincuts.
\newblock In \emph{Proceedings of the Twenty-first International Conference on
  Machine Learning}, ICML '04, pages 13--, New York, NY, USA, 2004. ACM.
\newblock ISBN 1-58113-838-5.
\newblock \doi{10.1145/1015330.1015429}.
\newblock URL \url{http://doi.acm.org/10.1145/1015330.1015429}.

\bibitem[Goodfellow et~al.(2014)Goodfellow, Shlens, and
  Szegedy]{goodfellow2014explaining}
I.~J. Goodfellow, J.~Shlens, and C.~Szegedy.
\newblock Explaining and harnessing adversarial examples.
\newblock \emph{arXiv preprint arXiv:1412.6572}, 2014.

\bibitem[Grandvalet and Bengio(2005)]{grandvalet2005semi}
Y.~Grandvalet and Y.~Bengio.
\newblock Semi-supervised learning by entropy minimization.
\newblock In \emph{Advances in neural information processing systems}, pages
  529--536, 2005.

\bibitem[He et~al.(2016)He, Zhang, Ren, and Sun]{he2016deep}
K.~He, X.~Zhang, S.~Ren, and J.~Sun.
\newblock Deep residual learning for image recognition.
\newblock In \emph{Proceedings of the IEEE conference on computer vision and
  pattern recognition}, pages 770--778, 2016.

\bibitem[Laine and Aila(2016)]{laine2016temporal}
S.~Laine and T.~Aila.
\newblock Temporal ensembling for semi-supervised learning.
\newblock \emph{arXiv preprint arXiv:1610.02242}, 2016.

\bibitem[Lee(2013)]{lee2013pseudo}
D.-H. Lee.
\newblock Pseudo-label: The simple and efficient semi-supervised learning
  method for deep neural networks.
\newblock In \emph{Workshop on Challenges in Representation Learning, ICML},
  volume~3, page~2, 2013.

\bibitem[Luo et~al.(2018)Luo, Zhu, Li, Ren, and Zhang]{luo2018smooth}
Y.~Luo, J.~Zhu, M.~Li, Y.~Ren, and B.~Zhang.
\newblock Smooth neighbors on teacher graphs for semi-supervised learning.
\newblock In \emph{Proceedings of the IEEE Conference on Computer Vision and
  Pattern Recognition}, pages 8896--8905, 2018.

\bibitem[Mitchell(1999)]{mitchell1999consumer}
V.-W. Mitchell.
\newblock Consumer perceived risk: conceptualisations and models.
\newblock \emph{European Journal of marketing}, 33\penalty0 (1/2):\penalty0
  163--195, 1999.

\bibitem[Miyato et~al.(2018)Miyato, Maeda, Ishii, and
  Koyama]{miyato2018virtual}
T.~Miyato, S.-i. Maeda, S.~Ishii, and M.~Koyama.
\newblock Virtual adversarial training: a regularization method for supervised
  and semi-supervised learning.
\newblock \emph{IEEE transactions on pattern analysis and machine
  intelligence}, 2018.

\bibitem[Oliver et~al.(2018)Oliver, Odena, Raffel, Cubuk, and
  Goodfellow]{oliver2018realistic}
A.~Oliver, A.~Odena, C.~A. Raffel, E.~D. Cubuk, and I.~Goodfellow.
\newblock Realistic evaluation of deep semi-supervised learning algorithms.
\newblock In \emph{Advances in Neural Information Processing Systems}, pages
  3239--3250, 2018.

\bibitem[Rasmus et~al.(2015)Rasmus, Berglund, Honkala, Valpola, and
  Raiko]{rasmus2015semi}
A.~Rasmus, M.~Berglund, M.~Honkala, H.~Valpola, and T.~Raiko.
\newblock Semi-supervised learning with ladder networks.
\newblock In \emph{Advances in neural information processing systems}, pages
  3546--3554, 2015.

\bibitem[Reed et~al.(1992)Reed, Oh, and Marks]{reed1992regularization}
R.~Reed, S.~Oh, and R.~Marks.
\newblock Regularization using jittered training data.
\newblock In \emph{[Proceedings 1992] IJCNN International Joint Conference on
  Neural Networks}, volume~3, pages 147--152. IEEE, 1992.

\bibitem[Riloff and Wiebe(2003)]{riloff2003learning}
E.~Riloff and J.~Wiebe.
\newblock Learning extraction patterns for subjective expressions.
\newblock In \emph{Proceedings of the 2003 conference on Empirical methods in
  natural language processing}, 2003.

\bibitem[Rosenberg et~al.(2005)Rosenberg, Hebert, and
  Schneiderman]{rosenberg2005semi}
C.~Rosenberg, M.~Hebert, and H.~Schneiderman.
\newblock Semi-supervised self-training of object detection models.
\newblock 2005.

\bibitem[Sajjadi et~al.(2016)Sajjadi, Javanmardi, and
  Tasdizen]{sajjadi2016regularization}
M.~Sajjadi, M.~Javanmardi, and T.~Tasdizen.
\newblock Regularization with stochastic transformations and perturbations for
  deep semi-supervised learning.
\newblock In \emph{Advances in Neural Information Processing Systems}, pages
  1163--1171, 2016.

\bibitem[Sietsma and Dow(1991)]{sietsma1991creating}
J.~Sietsma and R.~J. Dow.
\newblock Creating artificial neural networks that generalize.
\newblock \emph{Neural networks}, 4\penalty0 (1):\penalty0 67--79, 1991.

\bibitem[Srivastava et~al.(2014)Srivastava, Hinton, Krizhevsky, Sutskever, and
  Salakhutdinov]{srivastava2014dropout}
N.~Srivastava, G.~Hinton, A.~Krizhevsky, I.~Sutskever, and R.~Salakhutdinov.
\newblock Dropout: a simple way to prevent neural networks from overfitting.
\newblock \emph{The Journal of Machine Learning Research}, 15\penalty0
  (1):\penalty0 1929--1958, 2014.

\bibitem[Tarvainen and Valpola(2017)]{tarvainen2017mean}
A.~Tarvainen and H.~Valpola.
\newblock Mean teachers are better role models: Weight-averaged consistency
  targets improve semi-supervised deep learning results.
\newblock In \emph{Advances in neural information processing systems}, pages
  1195--1204, 2017.

\bibitem[Weston et~al.(2012)Weston, Ratle, Mobahi, and
  Collobert]{weston2012deep}
J.~Weston, F.~Ratle, H.~Mobahi, and R.~Collobert.
\newblock Deep learning via semi-supervised embedding.
\newblock In \emph{Neural Networks: Tricks of the Trade}, pages 639--655.
  Springer, 2012.

\bibitem[Yang et~al.(2016)Yang, Cohen, and Salakhutdinov]{yang2016revisiting}
Z.~Yang, W.~W. Cohen, and R.~Salakhutdinov.
\newblock Revisiting semi-supervised learning with graph embeddings.
\newblock \emph{arXiv preprint arXiv:1603.08861}, 2016.

\bibitem[Yarowsky(1995)]{yarowsky1995unsupervised}
D.~Yarowsky.
\newblock Unsupervised word sense disambiguation rivaling supervised methods.
\newblock In \emph{33rd annual meeting of the association for computational
  linguistics}, 1995.

\bibitem[Zhu and Ghahramani(2002)]{zhu2002learning}
X.~Zhu and Z.~Ghahramani.
\newblock Learning from labeled and unlabeled data with label propagation.
\newblock Technical report, Citeseer, 2002.

\bibitem[Zhu et~al.(2003)Zhu, Ghahramani, and Lafferty]{zhu2003semi}
X.~Zhu, Z.~Ghahramani, and J.~D. Lafferty.
\newblock Semi-supervised learning using gaussian fields and harmonic
  functions.
\newblock In \emph{Proceedings of the 20th International conference on Machine
  learning (ICML-03)}, pages 912--919, 2003.

\bibitem[Zhu(2005)]{zhu2005semi}
X.~J. Zhu.
\newblock Semi-supervised learning literature survey.
\newblock Technical report, University of Wisconsin-Madison Department of
  Computer Sciences, 2005.

\end{thebibliography}

\end{document}